\def\year{2018}\relax
\documentclass[letterpaper]{article} 
\usepackage{aaai18}  
\usepackage{times}  
\usepackage{helvet}  
\usepackage{courier}  
\usepackage{url}  
\usepackage{hyperref}

\usepackage{todonotes}
\usepackage{tablefootnote}
\usepackage{adjustbox}
\usepackage{threeparttable}
\usepackage{pdfpages}
\usepackage{comment}

\usepackage{subcaption}
\usepackage{amsmath}

\nocopyright

\usepackage{graphicx}  
\begin{document}
%

\title{HyperENTM: Evolving Scalable Neural Turing Machines through HyperNEAT}
\author{Jakob Merrild \and Mikkel Angaju Rasmussen \and Sebastian Risi\\IT University of Copenhagen, Denmark\\
\{jmer, mang, sebr\}@itu.dk
}

\maketitle

\begin{abstract}
Recent developments within memory-augmented neural networks have solved sequential problems requiring long-term memory, which are intractable for traditional neural networks. However, current approaches still struggle to scale to large memory sizes and sequence lengths. In this  paper we show how access to memory can be encoded geometrically through a  HyperNEAT-based Neural Turing Machine (\textit{HyperENTM}). We demonstrate that using the indirect  HyperNEAT encoding  allows for training on small memory vectors in a bit-vector copy task and then applying the knowledge  gained from such training to speed up training on larger size memory vectors. Additionally, we demonstrate that in some instances, networks trained to copy bit-vectors of size 9 can be scaled to sizes of 1,000 \textit{without further training}. While the task in this paper is simple, these results could open up the problems amendable to networks with external memories to problems with larger memory vectors and theoretically unbounded memory sizes.   
\end{abstract}

\section{Introduction}
Memory-augmented neural networks are a recent improvement on artificial neural networks (ANNs) that allow them to solve complex sequential tasks requiring long-term memory \cite{sukhbaatar2015end,graves2016hybrid,DBLP:journals/corr/GravesWD14} . Here we are particularly interested in Neural Turing Machines (NTM) \cite{DBLP:journals/corr/GravesWD14}, which allow a 
network to use an external memory tape to read and write information during execution. This improvement enables the ANN to be trained 
in fewer iterations, for certain tasks, and also allows the network to change 
behaviour on the fly. 

However, scaling to large memory sizes and sequence length is still challenging. Additionally, current algorithms have difficulties \textit{extrapolating} information learned on smaller problem sizes to larger once, thereby bootstrapping from it. For example, in the copy tasks introduced by Graves et al.~\shortcite{DBLP:journals/corr/GravesWD14} the goal is to store and later recall a sequence of bit vectors of a specific size. It would be desirable that a network trained on a certain bit vector size (e.g.\ 8 bits) would be able to scale to larger bit vector sizes without further training. However, current machine learning approaches often cannot transfer such knowledge. 

Recently, Greve et al.~\shortcite{Greve:2016:ENT:2908812.2908930} introduced an \textit{evolvable} version of the NTM (ENTM),  which did not rely on differentiability and offered some unique advantages. First, in addition to the network’s weights, the optimal neural architecture
can be learned at the same time. Second, a hard memory attention mechanism is directly
supported and the complete memory does not need to be accessed each time step. Third, a growing and theoretically infinite memory is now possible. Additionally, in contrast to the original NTM, the network was able to perfectly scale to very long sequence lengths. However, because it employed the direct genetic encoding NEAT, which means that every parameter of the network is described separately in its genotype, the approach had problems scaling to copy tasks with vectors of more than 8 bits. 

To overcome these challenges, in this paper we combine the ENTM with the indirect Hypercube-based NeuroEvolution
of Augmenting Topologies (HyperNEAT) encoding \cite{stanley2009hypercube}. HyperNEAT  provided a new perspective on evolving ANNs by
showing that the pattern of weights across the connectivity
of an ANN can be generated as a function of its geometry.
HyperNEAT employs an indirect encoding called compositional
pattern producing networks (CPPNs), which can
compactly encode patterns with regularities such as symmetry,
repetition, and repetition with variation \cite{stanley:gpem07}.
HyperNEAT exposed the fact that neuroevolution benefits
from neurons that exist at locations within the space of the
brain and that by placing neurons at locations, evolution can
exploit topography (as opposed to just topology), which makes
it possible to correlate the geometry of sensors with the
geometry of the brain. While lacking in many ANNs, such geometry is
a critical facet of natural brains 
\cite{sporns2002network}.
This insight allowed large ANNs with regularities in connectivity
to evolve for high-dimensional problems. 

In the new approach introduced in this paper, called \textit{HyperENTM}, an evolved neural network generates the weights of a main model, \textit{including how it connects to the external memory component.} Because HyperNEAT can learn the geometry of how the network should be connected to the external memory, it is possible to train a CPPN on a small bit vector sizes and then scale to larger bit vector sizes \textit{without further training}. 

While the task in this paper is simple it shows -- for the first time -- that access to an external memory can be indirectly encoded, an insight that could directly benefit indirectly encoded HyperNetworks training through gradient descent \cite{ha2016hypernetworks} and could be applied to more complex problem by employing recent advances in Evolutionary Strategies \cite{salimans2017evolution}.

\section{Backgorund}
This section reviews NEAT, HyperNEAT, and Evolvable Neural Turing Machines, which are foundational to the approach introduced in this paper.

\subsection{Neuroevolution of Augmenting Topologies}

The HyperNEAT method that enables learning from geometry in this paper
is an extension of the original NEAT algorithm that evolves ANNs through a \emph{direct} encoding \cite{stanley:ec02,stanley:jair04}. It starts with a
population of simple neural networks and then
\emph{complexifies} them over generations by adding new nodes and
con\-nections through mutation. By evolving networks in this way, the
topology of the network does not need to be known a priori; NEAT
searches through increasingly complex networks to find a suitable
level of complexity.

The important feature of NEAT for the purpose of this paper is that it
evolves \emph{both} the topology and weights of a 
network. Because it starts simply and gradually adds complexity, it
tends to find a solution network close to the minimal necessary
size.  The next section reviews the HyperNEAT extension to NEAT that is itself extended in this paper.
\subsection{HyperNEAT}
\label{sec:cppn}
In direct encodings like NEAT, each part of the solution's representation maps to a single piece of structure in the final solution \cite{floreano08neuroevolution}.
The significant disadvantage of this approach is that even when different parts of the solution are similar, they must be encoded and therefore discovered separately.  Thus this paper employs an \emph{indirect} encoding instead, which means that the description of the solution is compressed such that information can be reused. Indirect encodings are powerful because they allow solutions to be represented as a \emph{pattern} of parameters, rather than requiring each parameter to be represented individually \cite{bongard:cec02,gauci2010,hornby:alife02,Stanley:alife03}.
HyperNEAT, reviewed in this section, is an indirect encoding extension of NEAT that is proven in a number of challenging domains that require discovering regularities~\cite{clune:cec09,gauci2010,stanley2009hypercube}. 
For a full description of HyperNEAT see \cite{gauci2010}.

In HyperNEAT, NEAT is altered to evolve an indirect encoding called \emph{compositional pattern producing networks} (CPPNs \cite{stanley:gpem07}) \emph{instead} of ANNs.  CPPNs, which are also networks, are designed to encode \emph{compositions of functions}, wherein each function in the composition loosely corresponds to a useful regularity.

The appeal of this encoding is that it allows spatial patterns to be represented as networks of simple functions (i.e.\ CPPNs), which means that NEAT can evolve CPPNs just like ANNs.  CPPNs are similar to ANNs, but they rely on more than one activation function (each representing a common regularity) and are an abstraction of biological development rather than of brains. 
The indirect CPPN encoding can compactly encode patterns with regularities such as symmetry, repetition, and repetition with variation \cite{stanley:gpem07}. For example, simply by including a Gaussian function, which is symmetric, the output pattern can become
symmetric. A periodic function such as sine creates segmentation through repetition. Most importantly,
\emph{repetition with variation} (e.g.\ such as the fingers of the human hand) is easily discovered by combining regular
coordinate frames (e.g.\ sine and Gaussian) with irregular ones (e.g.\ the asymmetric x-axis). 
The potential for CPPNs to represent patterns with motifs reminiscent of patterns in natural organisms has been demonstrated in several studies \cite{secretan:chi08,stanley:gpem07}.   

The main idea in HyperNEAT is that CPPNs can naturally encode \emph{connectivity patterns} \cite{gauci2010,stanley2009hypercube}.  That way, NEAT can evolve CPPNs that represent large-scale ANNs with their own symmetries and regularities. 

\begin{figure}
 \begin{center}
\includegraphics[width=3in]{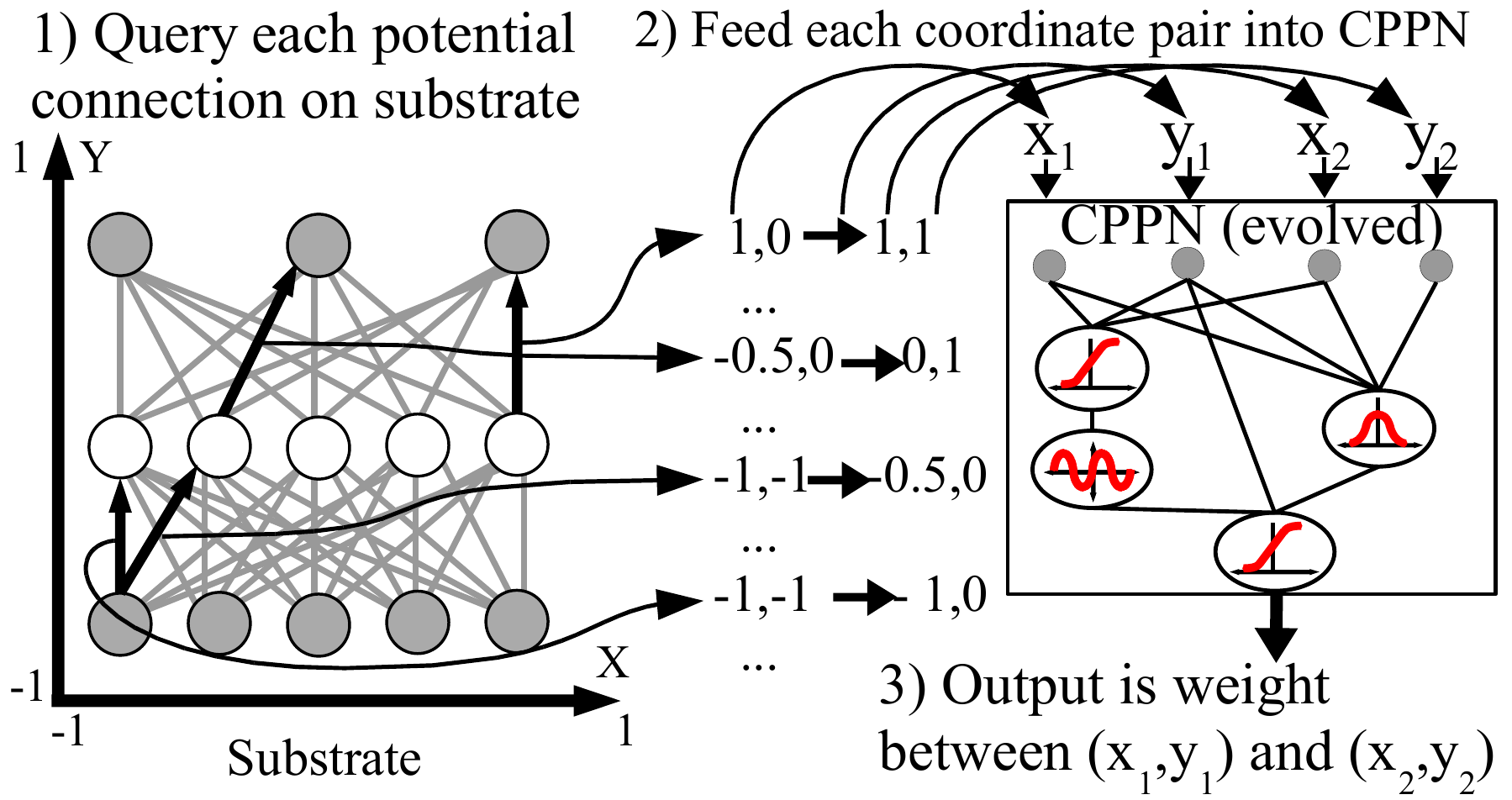}
\vspace{-0.1in}
\end{center}
 \caption{Hypercube-based Geometric Connectivity Pattern Interpretation.  \normalfont
  {
A collection nodes, called the \emph{substrate}, is assigned coordinates
that range from $-1$ to $1$ in all dimensions.
(1) Every potential connection in the substrate is queried to
determine its presence and weight; the dark directed
lines in the substrate depicted in the figure
represent a sample of connections that are queried.
(2) Internally, the CPPN (which is evolved) is a graph that determines which activation functions are connected.  
As in an ANN, the connections are weighted such that the output of a function is multiplied 
by the weight of its outgoing connection. For each query, the CPPN takes as input the positions of 
the two endpoints and (3) outputs the weight of the connection between them.
Thus, CPPNs can produce
regular patterns of connections in space.
}}
\label{fg:connective_interp}
\vspace{-0.15in}
\end{figure}
Formally, CPPNs are \emph{functions} of geometry (i.e.\ locations in space) that output connectivity patterns whose nodes are situated in $n$ dimensions, where $n$ is the number of dimensions in a Cartesian space. Consider a CPPN that takes four inputs labeled $x_1,y_1,x_2,$ and $y_2$; this point in four-dimensional space \emph{also} denotes the connection between the two-dimensional points $(x_1,y_1)$ and $(x_2,y_2)$, and the output of the CPPN for that input thereby represents the weight of that connection (Figure~\ref{fg:connective_interp}).  By querying every possible connection among a pre-chosen set of points in this manner, a CPPN can produce an ANN, wherein each queried point is a neuron position.  Because the connections are produced by a function of their endpoints, the final structure is produced with \emph{knowledge} of its geometry.  

In HyperNEAT, the experimenter defines both the location and role (i.e.\ hidden,
input, or output) of each such node.  As a rule of thumb, nodes are
placed on the substrate to reflect the geometry of the task
\cite{clune:cec09,stanley2009hypercube}. That
way, the connectivity of the substrate is a function of the task
structure. How to integrate this setup with an ANN that has an external memory component is an open question, which this paper tries to address. 


\subsection{Evolvable Neural Turing Machines (ENTM)}
\label{sec:entm}

Based on the principles behind the NTM, the recently introduced ENTM uses NEAT to learn the topology and weights of the ANN controller \cite{Greve:2016:ENT:2908812.2908930}. That way the topology of the network does not have to be defined a priori (as is the case in the original NTM setup) and the network can grow in response to the  complexity of the task. 
As demonstrated by Greve et al., the ENTM often finds compact network topologies to solve a particular task, thereby avoiding searching through unnecessarily high-dimensional spaces. Additionally, the ENTM was able to solve a complex continual learning problem \cite{luders2017continual}. 
Because the network does not have to be differentiable, it can use hard attention and shift mechanisms, allowing it to generalize perfectly to longer sequences in a copy task. Additionally, a dynamic, theoretically unlimited tape size is now possible.

The ENTM has a single combined read/write head. 
The network emits a write vector $w$ of size $M$, a write interpolation control input $i$, a content jump control input $j$, and three shift control inputs $s_l$, $s_0$, and $s_r$ (left shift, no shift, right shift). The size of the write vector $M$  determines the size of each memory location on the tape. The write interpolation component allows blending between the write vector and the current tape values at the write position, where $M_h(t)$ is the content of the tape at the current head location $h$, at time step $t$, $i_t$ is the write interpolation, and $w_t$ is the write vector, all at time step $t$:
  $M_h(t) = M_h(t-1) \cdot (1-i_t) + w_t \cdot i_t.$

The content jump determines if the head should be moved to the location in memory that most closely resembles the write vector. A content jump is performed if the value of the control input exceeds $0.5$. The similarity between write vector $w$ and memory vector $m$ is determined by:
    $s(w,m) = \frac{\sum^M_{i=1}|w_i-m_i|}{M}.$
At each time step $t$, the following actions are performed in order: (1) Record the write vector $w_t$ to the current head position $h$, interpolated with the existing content according to the write interpolation $i_t$. (2) If the content jump control input $j_t$ is greater than $0.5$, move the head to location on the tape most similar to the write vector $w_t$. (3) Shift the head one position left or right on the tape, or stay at the current location, according to the shift control inputs $s_l$, $s_0$, and $s_r$. (4) Read and return the tape values at the new head position.

\section{Approach: Hyper Neural Turing Machines (HyperENTM)}
In the HyperENTM the CPPN does not only determine the connections between the task related ANN inputs and outputs but also how the information coming from the memory is integrated into the network and how information is written back to memory.  Because HyperNEAT can learn the geometry of a task it should be able to learn the \textit{geometric pattern} in the information written to and read from memory. 

The following section details the HyperENTM approach on the copy task, which was first introduced by \cite{DBLP:journals/corr/GravesWD14}. In the copy task the network is asked to store and later 
recall a sequence of bit vectors. At the start of the task the network receives a 
special input, which denotes the start of the input phase. Afterwards, the 
network receives the sequence of bit-vectors, one at a time. Once the sequence 
has been fully passed to the network it receives another special input, 
signaling the end of the input phase and the start of the output phase. For 
any subsequent time steps the network does not receive any input.

In summary, the network has the following inputs: \emph{Start:} An input that is activated when the storing of 
		numbers should begin. \emph{Switch:} An input that is activated when the storing 
		should 
		stop and the network must start recalling the bit vectors instead. \emph{Bit-vector input:} Before the switch input has been 
		activated this input range is 
		activated with the bits that are to be recited later.  \emph{Memory read input:} The memory vector that the 
	TM read in the previous time step.

And following outputs: \emph{Bit-vector output:} The bit vector that the network 
		outputs to the environment. During the input phase this output is 
		ignored.  \emph{Memory write output:} The memory vector that should be 
		written to memory. \emph{TM controls:} TM specific control 
		outputs. Jump, 	interpolation, and three shift controls (left, stay, 
		and right).

\begin{figure*}	
	\begin{subfigure}[b]{.37\textwidth}\centering
		\includegraphics[width=0.9\textwidth]{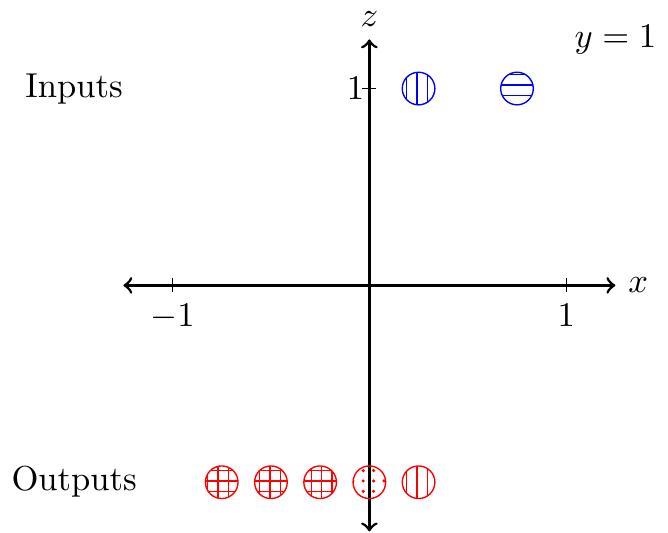}
		\caption{\label{fig:suby1}}	
	\end{subfigure}
	\begin{subfigure}[b]{.57\textwidth}\centering
	\includegraphics[width=0.9\textwidth]{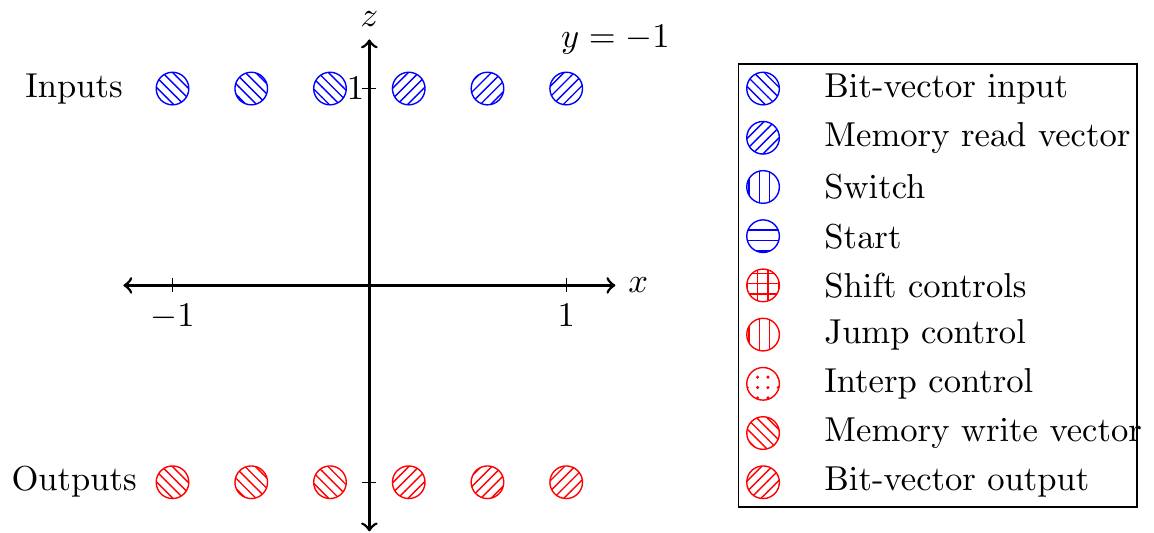}
		\caption{\label{fig:suby-1}}
	\end{subfigure}
	\caption[The substrate used for non-MSS experiments]{The HyperENTM substrate 
	for the copy task. All inputs are in $z = 1$ 
	and all outputs in $z = -1$. Figure (\subref{fig:suby1}) shows all nodes 
	in 
	$y = 1$ which is the start/switch inputs and the tm controls. 
	Notably the	$x$-coordinate is the same for the switch input and the jump 
	control output. 
	Figure (\subref{fig:suby-1}) shows the nodes in $y = -1$ which are the 
	bit-vector and memory vector input and outputs. Bit-vector input nodes share
	$x$-coordinates with memory vector write nodes, while memory vector read 
	nodes share $x$-coordinates with bit-vector output nodes.}
	\label{copysubstratecrowded}
\end{figure*}
\subsection{Copy Task Substrate}
\label{sec:substrates}
The substrate for the copy task is shown in Figure~\ref{copysubstratecrowded}. 
The substrate is  designed such that the bit-vector input nodes share $x$-coordinates with the  memory vector write nodes and vice versa with memory vector read nodes and  bit-vector output nodes.  Furthermore, the switch input  shares its $x$-coordinate with the jump output, thus encouraging the network to jump in memory when it should start reciting. In this paper, the size of the memory vector equals the bit vector size. Furthermore, none of the substrates contain 
hidden nodes as it has been shown that it is possible to solve even large 
versions of the problem without any hidden nodes~\cite{greve2016evolving}.

Following Verbancsics and Stanley~\shortcite{Verbancsics:2011:CCE:2001576.2001776}, in addition to the CPPN output that determines the weights of each connection, each CPPN has an additional step-function output, called the \textit{link-expression output} (LEO), which determines if a connections should be expressed. Potential connections are queried for each input on layers $y=1$ and $y=-1$ to each output on layers $y=1$ and $y=-1$. The number of inputs and outputs on the $y=-1$ layer (Figure~\ref{copysubstratecrowded}b) are scaled dependent on the size of the copy task bit vector. In the shown example the bit-vector size is three. Neurons are uniformly distributed in the $x$ interval $[-1.0, -0.2]$ for bit vector inputs and memory write vector and in the interval $[0.2, 1.0]$ for the memory read vector and bit vector output. 

The CPPN has an additional output that determines the bias values for each nodes in the substrate. These values are determine through node-centric CPPN queries (i.e.\ both source and target neuron positions $xyz$ are set to the location of the node whose bias should be determined).

\section{Experiments}
A total of three different approaches are evaluated on bit-sizes of 1, 3, 5, and 9. \textbf{HyperNEAT} is compared to the direct \textbf{NEAT} encoding, and a S\textbf{eeded HyperNEAT} treatment that starts evolution with a CPPN seed that encourages locality on both the $x$- and $y$-coordinates (Figure~\ref{fig:genomes}a). A similar locality seed has been shown useful in HyperNEAT to encourage the evolution of modular networks \cite{Verbancsics:2011:CCE:2001576.2001776}. This locality seed is then later adjusted by evolution (e.g.\ adding/removing nodes and connections and changing their weights).

The \textbf{fitness function} in this paper follows the one in the original ENTM paper \cite{Greve:2016:ENT:2908812.2908930}. During training the network is given a sequence of random bit 
vectors, between 1 and 10 vectors long, and asked to recite it. The network is tasked to do this with 50 random sequences and assigned a normalized fitness. The network is evaluation the bit-vectors recited by the network are compared to those given to it during the input phase; for every bit-vector the network is given a score based on how close the output from the network corresponding to a  specific bit was to the target. If the bit was within $0.25$ of the target, the network is awarded a fitness of the difference between the actual output and  the target output. Otherwise, the network is not awarded for that specific 
bit: $f = 1 - \frac{|x-x_t|}{0.25} $ if $|x-x_t| < 0.2$ and 0 otherwise.
The fitness for any given bit-vector is equal to the the sum of the fitness  for each individual bit, normalized to the length of the bit-vector. Similarly,  the fitness for a complete sequence is the sum of the fitness for each  bit-vector normalized to the length of the sequence. This results in a fitness score between 0 and 1. This fitness function rewards the network for gradually getting closer to the solution, but it does not actively reward the network for using the memory to store the inputs.

\subsection{Experimental Parameters}
For the NEAT experiments, offspring generation proportions are 50\% sexual (crossover) and
50\% asexual (mutation). Following \cite{luders2017continual}, we use 98.8\% synapse weight mutation probability, 9\% synapse addition probability, and 5\% synapse
removal probability. Node addition probability
is set to 0.05\%. The NEAT implementation SharpNEAT uses a complexity
regulation strategy for the evolutionary process, which has proven to be quite impactful on our results. A threshold defines how complex the networks in the population can be (here defined as the number of genes in the genome and set to 10 in our experiments), before the algorithm switches to a simplifying phase, where it gradually reduces complexity.

For the HyperNEAT experiments the following parameters are used. Elitism proportion is 2\%. Offspring generation proportions are 50\% sexual (crossover) and 50\% asexual (mutation). CPPN connection weights have a 98.8\% probability of being changed, a 1\% change of connection addition, and 0.1\% change of node addition and node deletion. The activation functions available to new neurons in the  CPPN are  Linear, Gaussian, Sigmoid, and Sine, each with a 25\% probability of being added.
Both NEAT and HyperNEAT experiments run with a population size of 500 for a maximum of 10,000 generations or until a solution is found. The code is available from: \url{https://github.com/kalanzai/ENTM_CSharpPort}. 

\begin{figure}[t]
	\centering
	\includegraphics[width=.47\textwidth]{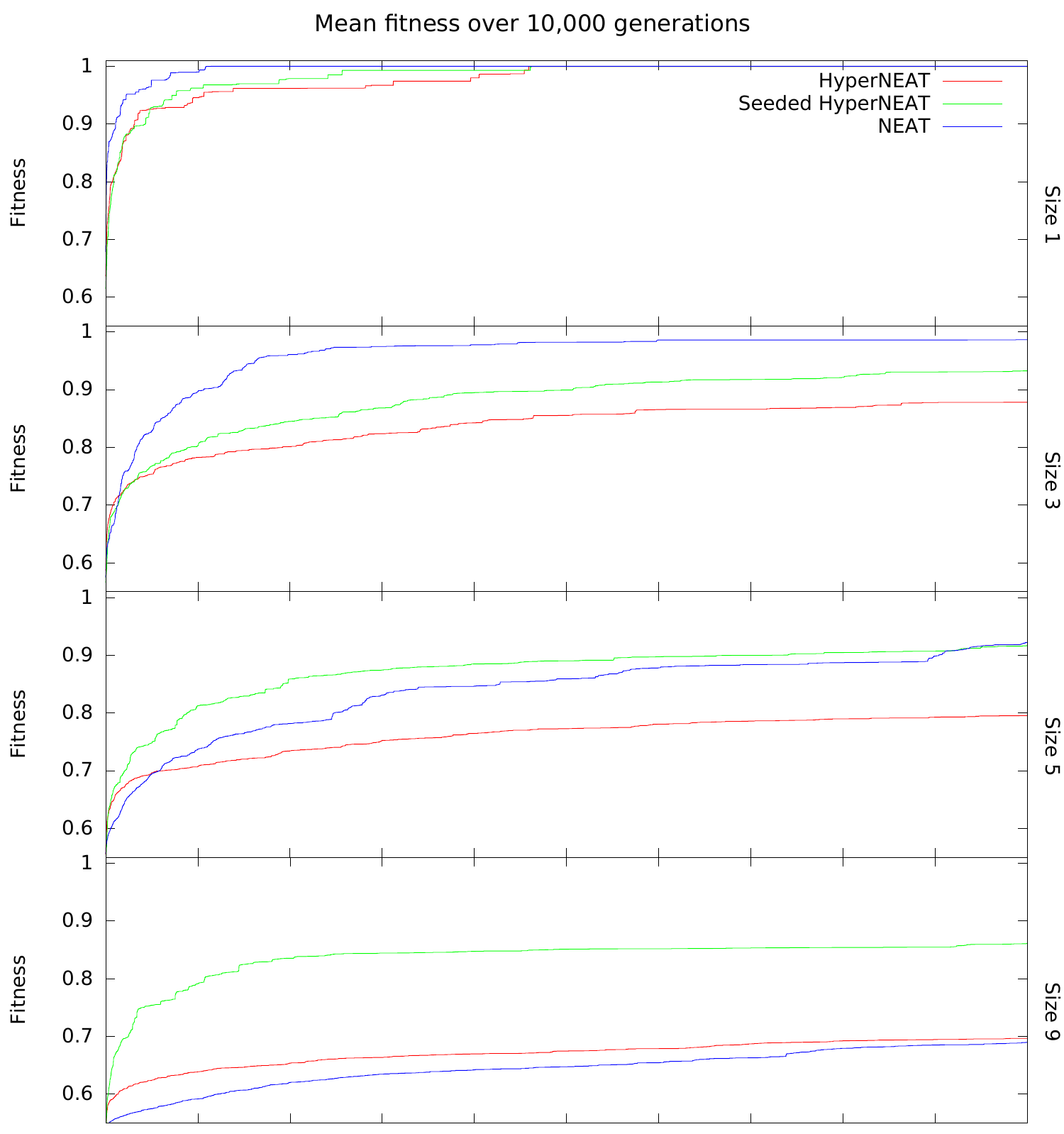}\vspace{-0.1in}
	\caption[Mean champion fitness over generations.]{Mean champion fitness for the different treatments and bit-sizes, averaged over 20 independent evolutionary runs.}
	\label{fig:fitnessGrowth}
\end{figure}



\section{Results}
\begin{figure}[h]
	\centering
	\includegraphics[width=.5\textwidth]{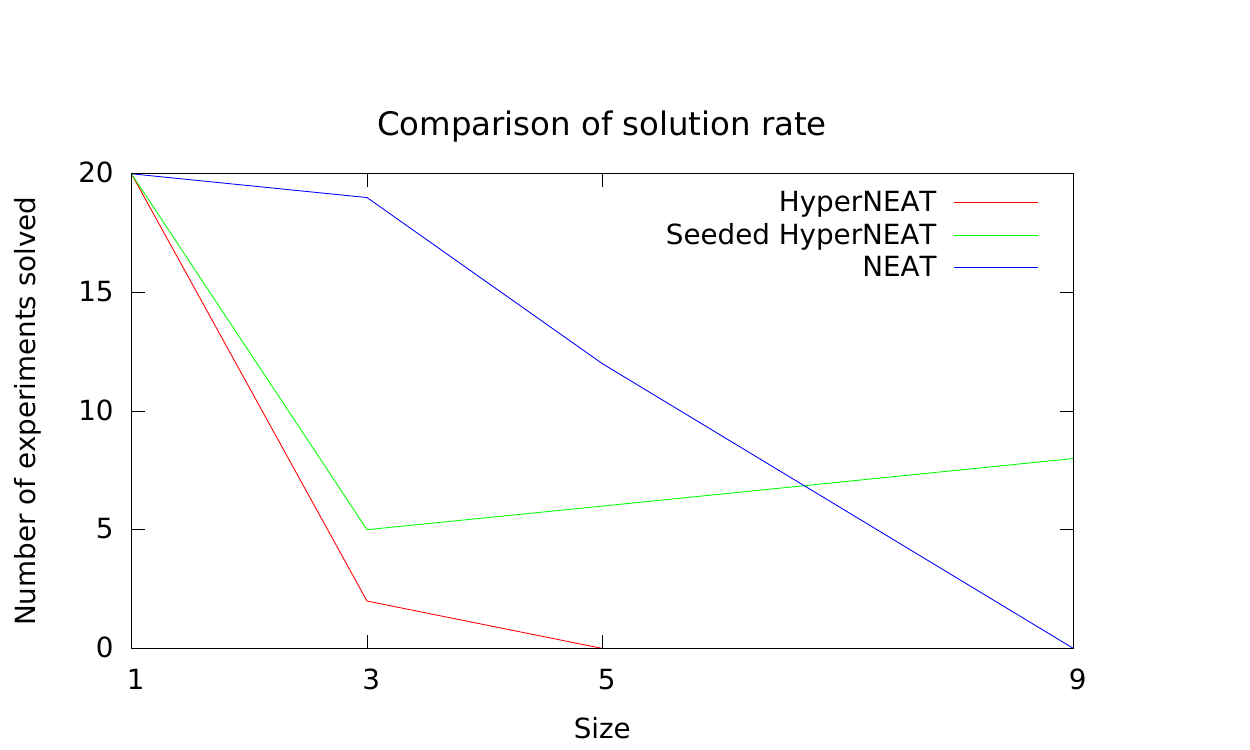}
	\caption{The number of solutions found by each configuration for the 
	four different bit vector sizes.}
	\label{fig:solutions}
\end{figure}

Figure~\ref{fig:fitnessGrowth} shows the mean champion fitness over 
$10,000$ generations for each of the different approaches and bit vector sizes. While NEAT performs best on smaller bit vectors, as the size of the vector grows to 9 bits, the seeded HyperNEAT variant outperforms both NEAT and HyperNEAT. 
The numbers of solutions found (i.e.\ networks that reach a training score $\geq$ 0.999) in regards to the bit vector size are shown in Figure~\ref{fig:solutions}. For bit size 1 all  approaches solve the problem equally 
well. However, as the size of the  bit vector is increased the 
configurations using HyperNEAT and locality seed performs best and the only method that is able to find any solution for size 9. 

\begin{figure}[t]
	\centering
	\includegraphics[width=.5\textwidth]{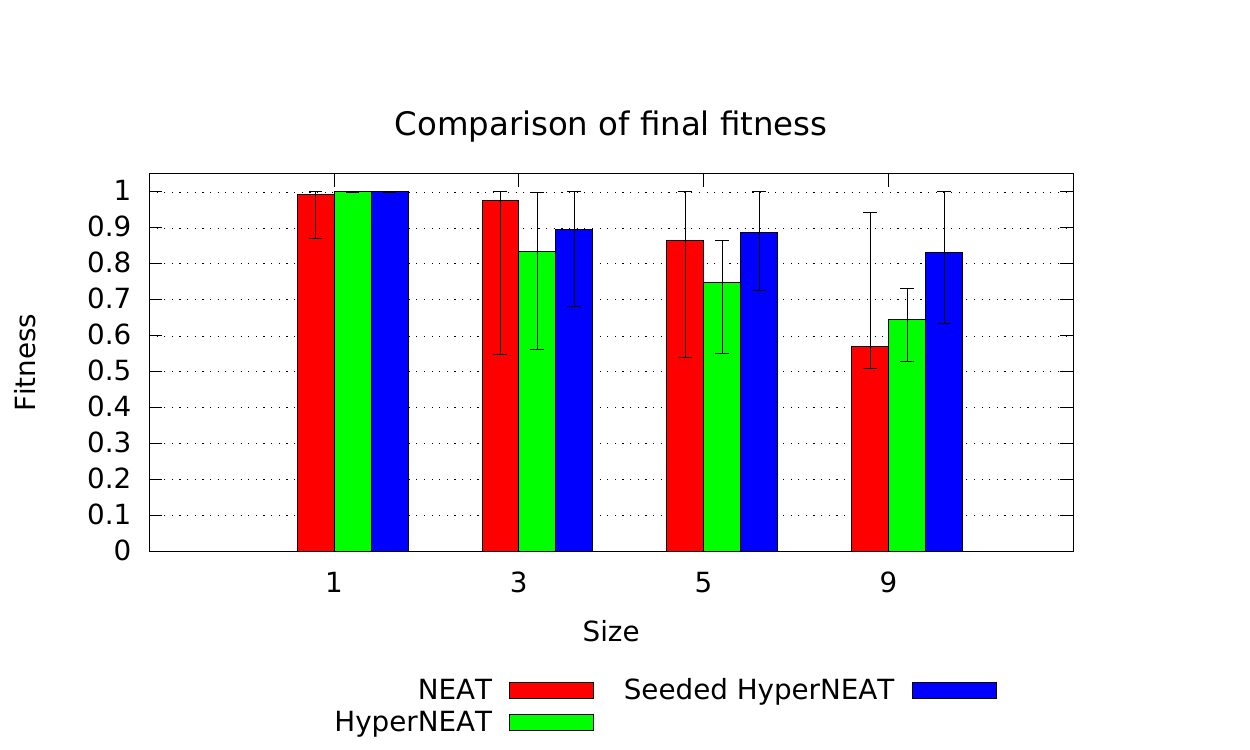}
	\caption[Genralisation score ...Mean champion fitness of final generation.]{The mean performance of the champion networks from the last generation. NEAT does well on smaller sizes, but degenerates quickly as the size goes up.}
	\label{fig:meanFitness}
\end{figure}

\textbf{Testing Performance.} To determine how well the champions from the last generation generalize, they were tested on $100$ random 
bit-vector sequences of a random lengths between 1 and 10 (Figure~\ref{fig:meanFitness}). 
On sizes 1 and 5 there is no statistical difference between either 
treatment (following a two-tailed Mann-Whitney U test). 
On size 3, NEAT performs significantly better than the seeded HyperNEAT ($p < .00001$).  Finally, seeded HyperNEAT performs  significantly 
better than NEAT on size 9 ($p < .00001 $). The main conclusions are that (1) while NEAT performs best on smaller bit vectors it degrades rapidly with increased bit sizes, and (2) the seeded HyperNEAT variant is able to scale to  larger sizes while maintaining performance better. 

        
\textbf{Generalizing to longer sequences. }  We also tested how many of the solutions, which were trained on bit sequences of length 10, generalize to sequences of length 100. The training and generalisation results are summarized in Table~\ref{tab:solutions}, which shows the number of solutions for each of the 
three approaches, how many of those solutions generalized to sequences of  length 100, and the average number of generations it took to find a solution. For all three methods, most solutions generalize perfectly to sequences that are longer than the sequences encountered during training.
\begin{table}
	\centering
	\caption{Generalisation Results. Shown are the number of solutions, the number of solutions that generalize, together with the average number of generations it took to find a solution and standard deviation.}
	\label{tab:solutions}
	\begin{threeparttable}
	\begin{adjustbox}{max width=\textwidth}
	\begin{tabular}{|c|c|c|c|c|c|}
		\hline 
		 & Size & \#sol. & \#gen & 
		gens. & sd.\\ 
		\hline 
		Seeded & 1 & 20 &20 & 1055.8 & 1147.4 \\
		HyperNEAT & 3 & 5 & 5 & 4454.8 & 3395.8 \\
		& 5 & 6& 5 & 2695.5 &2666.5 \\
		& 9 & 8 & 5 & 1523.25 & 2004.1\\\hline 
		HyperNEAT & 1 & 20 &20 &1481.45 & 1670.8\\
		& 3 & 2 & 2 & 4395.5 & 388.2\\
		& 5 & 0 & 0 & N/A & N/A \\
		& 9 & 0 & 0 & N/A & N/A \\\hline 
		NEAT & 1 & 20 & 19 &281 &336.3 \\
		& 3 & 19 & 19 &2140.5 &1594.4 \\
		& 5 &12 & 11 & 3213.4&2254.2 \\
		& 9 & 0& 0& N/A&N/A \\\hline 
	\end{tabular}	
	\end{adjustbox}
	\end{threeparttable}
\end{table}

\subsection{Transfer Learning}

To test the scalability of the Seeded HyperNEAT solutions, champion genomes from runs which found a solution for a given size 
were used as a seed for evolutionary runs of higher sizes. The specific runs 
and which seeds were used can be seen in Table~\ref{tab:scalingResults},  which 
also contains the number of solutions found, how many 
solutions generalized, and the average number of generations needed to find the 
solutions. Seeds denoted $X\rightarrow Y$ refer to champion genomes from a run 
of size $Y$ which was seeded with a champion from a run of size $X$.
\begin{table}[t]
	\caption{HyperNEAT Transfer Learning}
	\label{tab:scalingResults}
	\begin{adjustbox}{max width=\textwidth}
	\begin{threeparttable}
	\begin{tabular}{|c|c|c|c|c|c|c|}
		\hline 
		 Seed & Size & \#sol & \#general & 
		gens. 
		& 
		sd. 
		\\ 
		\hline 
		3 & 5 & 12/20 & 6 & 434.1 & 574.6 \\
		3 & 9 & 16/20 & 2 & 1150.3 & 2935.9 \\
		5 & 9 & 23/24 & 11 & 58.9 & 129.2 \\
		3$\rightarrow$5 & 9 & 23/24 & 8 & 588 & 1992\\
		 5 & 17 & 18/20 & 12 & 258.9 & 301.3 \\
		9 & 17 & 24/24 & 12 & 89.3 & 235.9 \\
		5$\rightarrow$9 & 17 & 20/20 & 17 & 36.25 & 47.5\\
		 9 & 33 & 23/24 & 17 &94 & 217	\\
		 9$\rightarrow$17 & 33 & 24/24 &19 &456.5 &2002.4 \\
		\hline
	\end{tabular}
	\end{threeparttable}
	\end{adjustbox}
\end{table}

\begin{figure}
	\includegraphics[width=.45\textwidth]{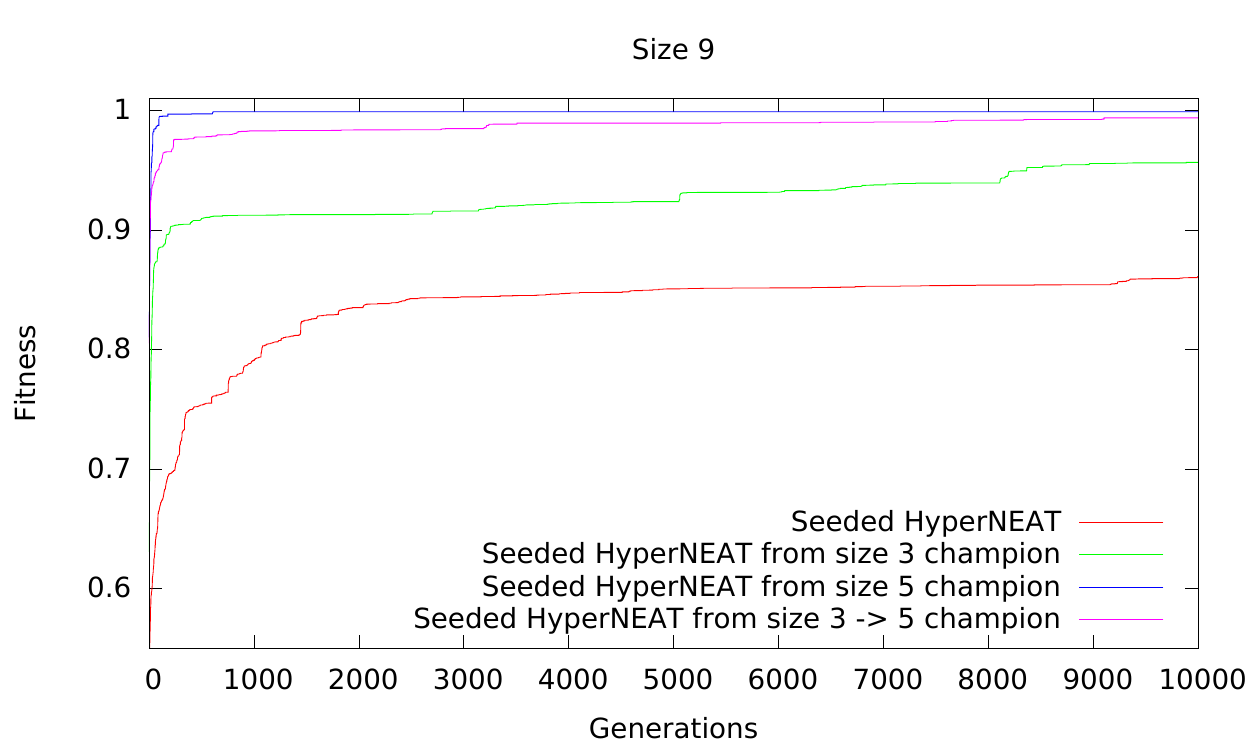}
	\caption{Comparison of the size 9 problem with normal locality seed and with champion seeds from a smaller size. }
	\label{fig:size9scaling}
\end{figure}

Because the number of solutions found varied between the different sizes (see Table~\ref{tab:solutions}), the scaling experiments were not run exactly 20  times. Instead, the number of runs was the smallest number above or equal to 20  which allowed for each champion to be seeded an equal number of times, e.g.\ if there were 6 solutions 24 runs were made; 4 runs with the champion from  each solution. Figure~\ref{fig:size9scaling} shows a comparison of HyperNEAT seeded with the locality seed and seeded with champion genomes of smaller sizes on the size 9 problem. HyperNEAT yielded significantly better results when seeded with size 5 and 3$\rightarrow$5 champions compared to starting with the locality seed ($p < .001$), but not when seeded with the champion from size 3.

\subsection{Scaling without further training}
The champions from runs which found a solution were tested for scaling to larger bit-vector sizes without further evolutionary training (i.e.\  new input and output nodes are created and queried by the CPPN; see copy task substrate section for details). 
Each genome was tested on 50 sequences of 100 random  bit-vectors of size 1,000.  Some of the  champions found using only the LEO size 9 configuration scaled perfectly  to a bit-size of 1,000 without further training, as seen in Table~\ref{tab:scale_no_evolve}.   

\begin{table}
	\centering
	\caption{Scaling using LEO without further evolution}
	\label{tab:scale_no_evolve}
	\begin{threeparttable}
	\begin{tabular}{|l | c | c |}
		\hline
		Size & \# of champions & \# which scaled to 1000 \\
		\hline
		9 & 8 & 2 \\
		\hline
		$9\rightarrow17$ & 24 & 7\tnote{$\dagger$} \\
		\hline
	\end{tabular}
	\begin{tablenotes}
		\item[$\dagger$] 6 of these can be traced back to the 2 champions from 
		size 9 which scaled perfectly.
	\end{tablenotes}
	\end{threeparttable}
\end{table}
\begin{figure}[t]
	\centering
	\begin{subfigure}[b]{.38\textwidth}
		\centering
		\includegraphics[width=\textwidth]{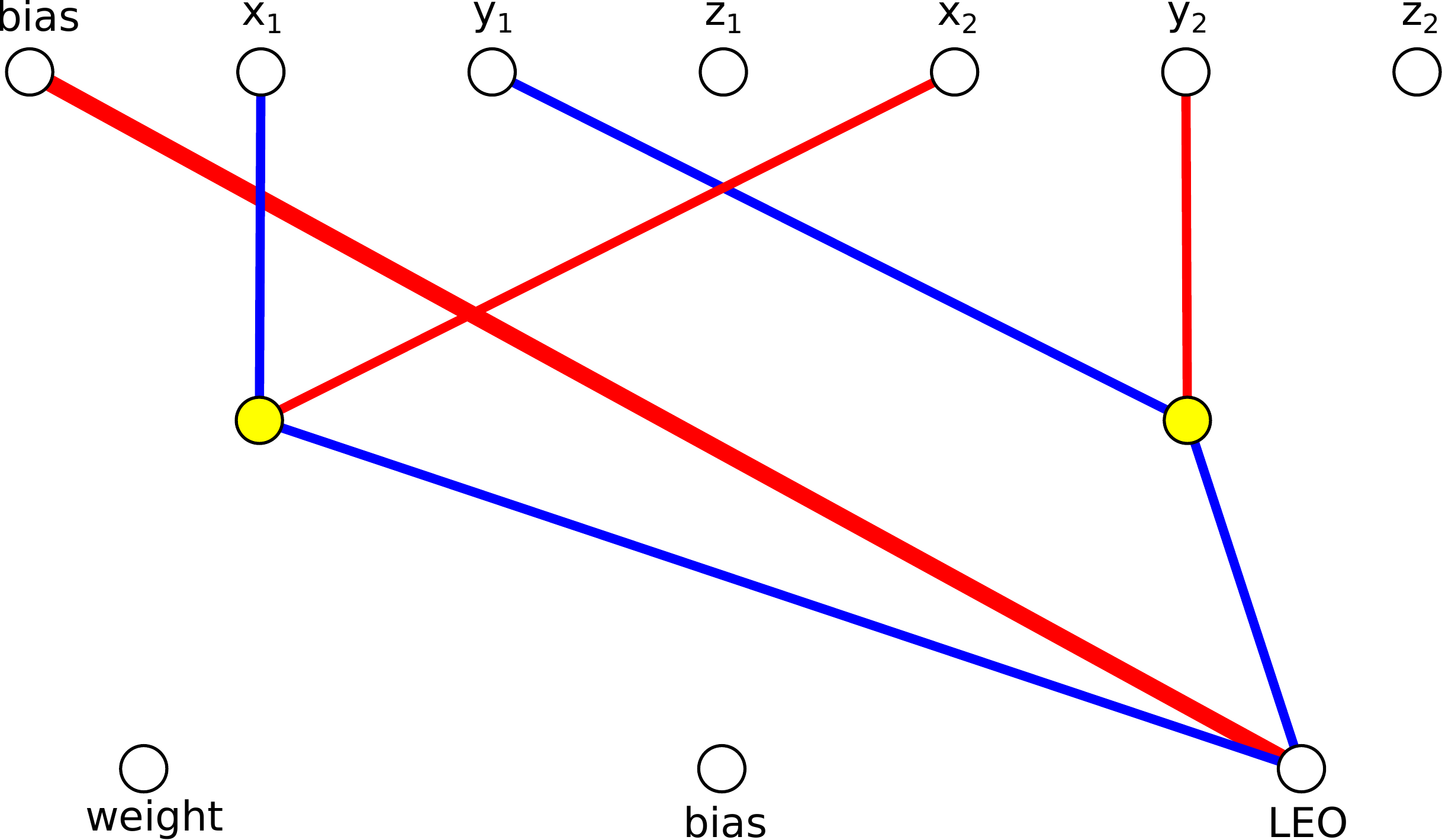}
		\caption{\label{fig:genome_seed}}	
	\end{subfigure}
	~
	\begin{subfigure}[b]{.38\textwidth}
	\centering
	\includegraphics[width=\textwidth]{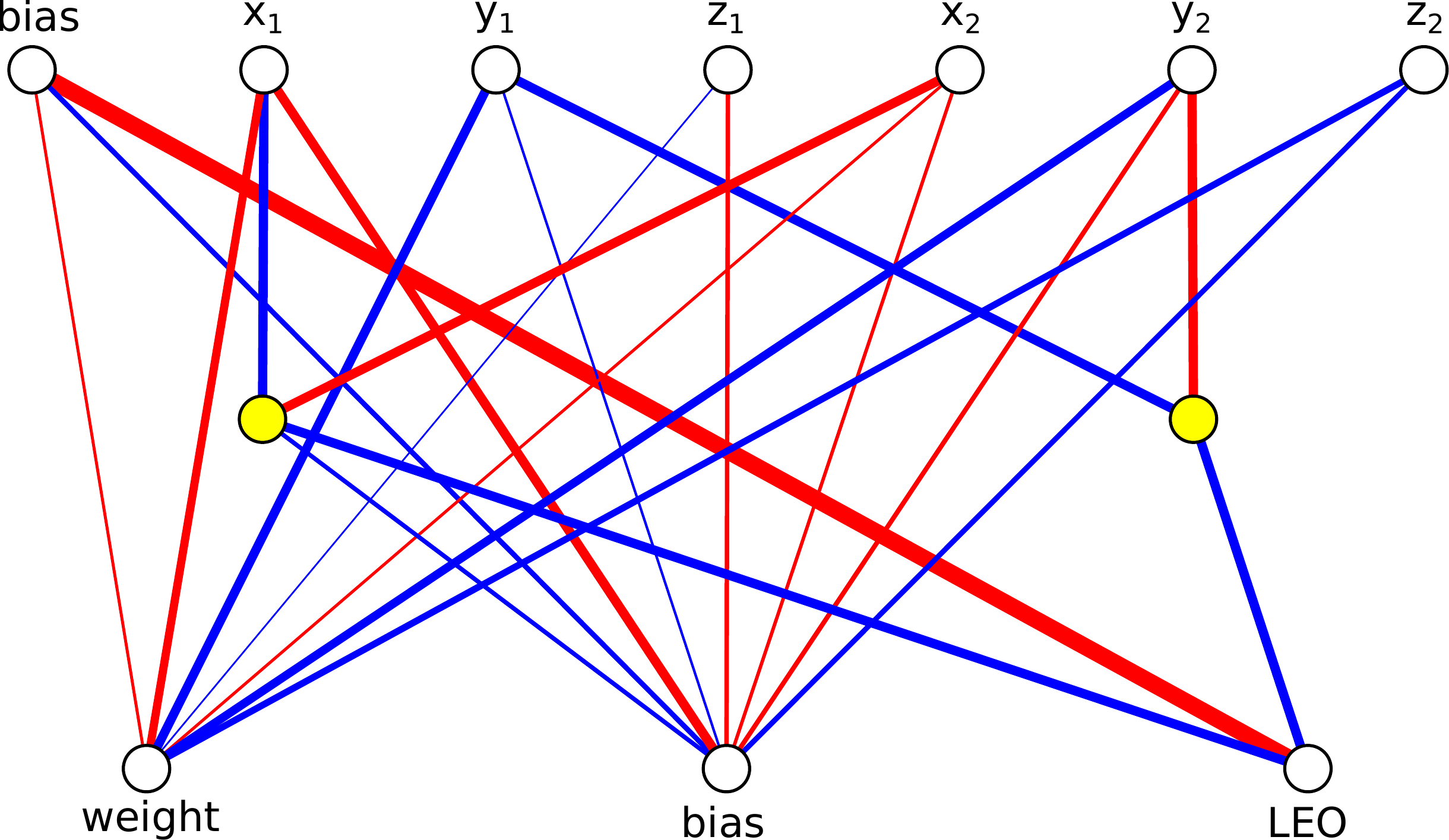}
	\caption{\label{fig:size9_champion_genome}}	
	\end{subfigure}
	\caption[Seed and Champion comparison]{(a) The CPPN seed that is 
	used to promote locality on 	the $x$ and $y$ axes. (b) A champion trained 
	on the size 9 problem, which was able to scale without 
	further evolution to size 1,000. Blue connections have a positive weight, while red connections have a negative weight. 
    }
	\label{fig:genomes}
\end{figure}
The main results is that it is possible to find CPPN that perfectly 
scale to any size. The fact that evolution with HyperNEAT performs 
significantly better, when seeded with a champion genome which solved a smaller 
size of the problem, together with the fact that evolution sometimes finds 
solutions which scale without further training, demonstrates that 
HyperNEAT can be used to scale the dimensionality of the bit-vector in 
the copy task domain.


\subsection{Solution Example}
Here we take a closer look at one of the champion 
genomes (trained on bit-vector size 9), which 
was able to scale perfectly to the size 1000 problem (Table~\ref{tab:scale_no_evolve}). 
Figure~\ref{fig:genomes} shows a visualization 
of the champion genome, as well as the locality seed from which it was evolved. The champion genome does resemble the seed but also evolved several additional connections that are necessary to solve the problem.

\begin{figure}[t]
	\centering
	\begin{subfigure}[b]{.45\textwidth}\centering
		\includegraphics[width=\textwidth]{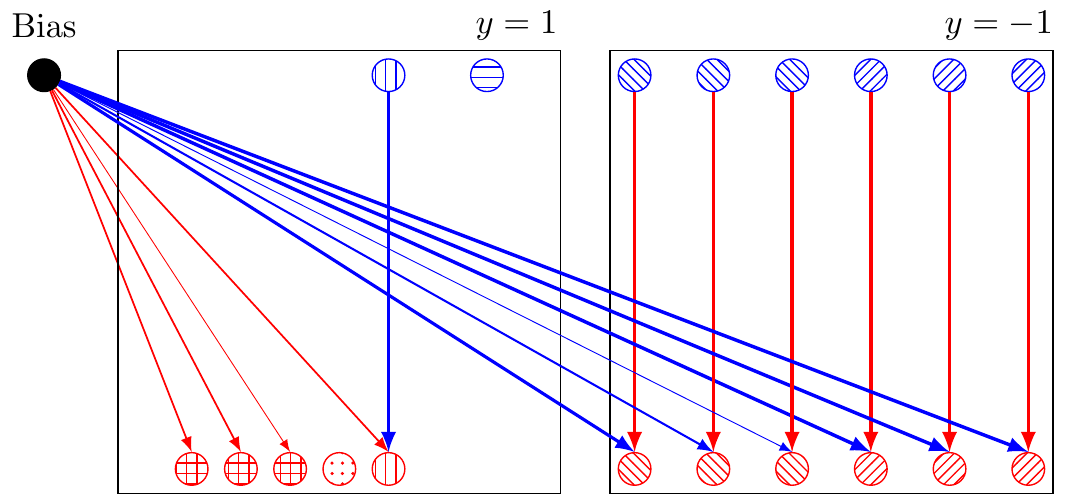}
		\caption{\label{fig:size3_phenome}}	
	\end{subfigure}
	~
	\begin{subfigure}[b]{.45\textwidth}\centering
		\includegraphics[width=\textwidth]{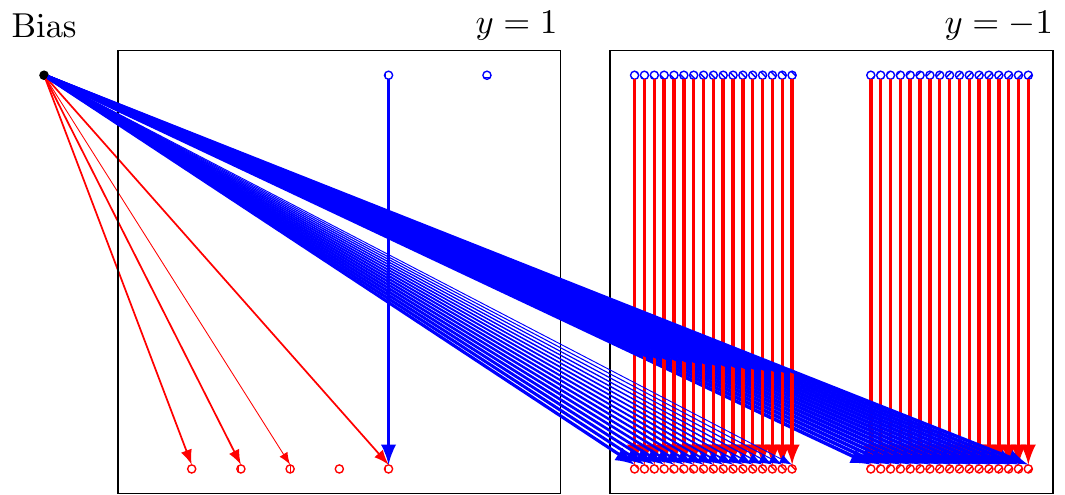}
		\caption{\label{fig:size17_phenome}}
	\end{subfigure}
	\caption[Phenome comparison]{Networks produced by the champion 
	CPPN (Figure~\ref{fig:size9_champion_genome}) for bit sizes  
	3 (\subref{fig:size3_phenome}) and  
	17 (\subref{fig:size17_phenome}). 
    }
	\label{fig:phenomes}
\end{figure}

Figure~\ref{fig:phenomes} shows two different ANNs generated by 
the the same CPPN, which is shown in Figure~\ref{fig:size9_champion_genome}.
It can be seen that for non-bias connections to be expressed, the source and 
destination nodes have to be located in the same position on both the $x$ position and  $y$ layer in the substrate. 
These results suggest that the locality encouraging seed works as intended.

\begin{figure*}
	\centering
	\begin{subfigure}[b]{.43\textwidth}
		\centering
		\includegraphics[width=\textwidth]{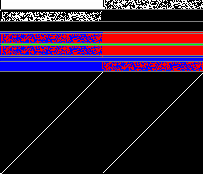}
		\caption{\label{fig:recording_size9}}
	\end{subfigure}
	\begin{subfigure}[b]{.43\textwidth}
		\centering
		\includegraphics[width=\textwidth]{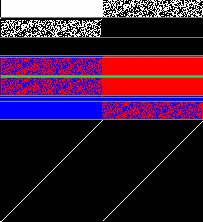}
		\caption{\label{fig:recording_size17}}
	\end{subfigure}
	\caption[Activity recordings example for size 9 and 17 ANNs]{Recordings 
	of the activities for bit-sizes 
		9 (\subref{fig:recording_size9})
		and 17 (\subref{fig:recording_size17}) networks. Both networks are produced by the same CPPN. See main text for details.}
	\label{fig:recordings}
\end{figure*}

To further demonstrate the scalability of this evolved CPPN, memory   usage for size 9 and 17 are shown 
in Figure~\ref{fig:recordings}. The output of the networks is shown at  
the top, followed by the input to the network. Next follows the difference 
between the given input and the output, i.e.\ how well the network recited the  sequence given to it. 
The fourth section of the recording 
shows a heat map of the fitness score based on the bit-vector in that 
position, where red indicates a high fitness while blue indicates a low 
fitness. 

Next follows the output from the network to the TM controller, followed 
by the \emph{interpolation} output, and what was written to the memory tape.  Finally, the recording shows the \emph{jump} output from the network, the three  different \emph{shift} outputs, what read from the tape, and the  position of  the read/write head. As the recording shows, the network which solves the problem jumps exactly 
once when the \emph{switch} input neuron is activated. 

\section{Conclusion}
This paper showed  that the indirect encoding HyperNEAT makes 
it feasible to train ENTMs with large memory vectors for a simple copy task, which would otherwise be infeasible to train with an direct encoding such as NEAT. Furthermore, starting with a CPPN seed that encouraged locality, it was possible 
to train solutions to the copy task that perfectly scale with the size of 
the bit vectors which should be memorized, without any further training. 
Lastly, we demonstrated that even solutions which do not scale perfectly can  be used to shorten the number of generations needed to evolve a solution for  bit-vectors of larger sizes. In the future it will be interesting to apply the approach to more complex domains, in which the geometry of the connectivity pattern to discover is more complex. Additionally, combining the presented approach with recent advances in Evolutionary Strategies \cite{salimans2017evolution}, which have been shown to allow evolution to scale to problems with extremely high-dimensionality is a promising next step.

\bibliographystyle{aaai}
\bibliography{references}


\end{document}